\documentclass{article}

\usepackage{times}  
\usepackage{hyperref}
\hypersetup{pdfstartview=FitH,pdfpagelayout=SinglePage}

\usepackage{graphicx}  
\usepackage{epstopdf}

\usepackage[affil-it]{authblk}

\usepackage{amsmath,amssymb}
\usepackage{url}
\usepackage{multirow}
\usepackage[usenames]{color}
\usepackage{times}
\usepackage{balance}
\usepackage{hyperref}
\usepackage[hyphenbreaks]{breakurl}
\usepackage{algorithm}
\usepackage{algorithmic}

\usepackage{subfigure}

\newcommand{\para}[1]{{\vspace{2pt} \bf \noindent #1 \hspace{6pt}}}

\newcommand\yzedit[1]{{\color{black} #1}}
\newcommand\zjedit[1]{{\color{black} #1}}

\newcommand \name{\texttt{Atia}}

\newenvironment{packed_itemize}{
\begin{list}{\labelitemi}{\leftmargin=0.8em}
  \setlength{\itemsep}{2pt}
  \setlength{\parskip}{0pt}
  \setlength{\parsep}{0pt}
  \setlength{\headsep}{0pt}
  \setlength{\topskip}{0pt}
  \setlength{\topmargin}{0pt}
  \setlength{\topsep}{0pt}
  \setlength{\partopsep}{0pt}
}{\end{list}}

\usepackage[accepted]{sysml2019}

\begin{document}

\twocolumn[
\sysmltitle{Addressing Training Bias via Automated Image Annotation}


\begin{sysmlauthorlist}
\sysmlauthor{Zhujun Xiao}{equal}
\sysmlauthor{Yanzi Zhu}{to}
\sysmlauthor{Yuxin Chen}{equal}
\sysmlauthor{Ben Y. Zhao}{equal}
\sysmlauthor{Junchen Jiang}{equal}
\sysmlauthor{Haitao Zheng}{equal}
\end{sysmlauthorlist}

\sysmlaffiliation{equal}{University of Chicago}
\sysmlaffiliation{to}{University of California, Santa Barbara}

\vskip 0.3in

\begin{abstract}
Build accurate DNN models requires training on large labeled, context
specific datasets, especially those matching the target scenario. 
We believe advances in wireless localization, working in unison with cameras,
  can produce automated annotation of targets on images and videos captured in the
  wild.   Using pedestrian and vehicle detection as examples, we demonstrate the feasibility,
  benefits, and challenges of an automatic image annotation system.  Our work  
  calls for new technical development on passive localization, mobile data
  analytics, and error-resilient ML models, as well as design issues in user privacy
  policies. 
\end{abstract}
]

\printAffiliationsAndNotice{} 

\section{Introduction}
\label{sec:intro}

Modern computer vision, in the forms of deep neural networks (DNNs), 
has promised to revolutionize many intelligent applications,  from  image and face recognition, 
to self-driving cars.
With dramatically increased expressiveness, DNNs can be tailored to produce inference 
results of unprecedented accuracy, when they are trained on sufficient dataset.

However, there is a roadblock to the success of DNNs in real applications: 
building accurate DNN models requires training on large 
{\em labeled, context-specific} 
datasets~\cite{kang2017noscope,andriluka2018posetrack,wang2018revisiting}.
DNNs trained on insufficiently labeled data have been reported to perform poorly
in multiple applications~\cite{zhu2012moredata}, from sign
language recognition~\cite{kim2017lexicon}, facial
recognition~\cite{kemelmacher2016megaface}, to urban vision 
applications for smart-cities~\cite{mallapuram2017smart}.

Building these training datasets, at least for urban vision tasks, is often
challenging for two reasons.  {\em First}, it requires manually labeling
images and videos, an extremely labor-intensive task.  While some have
proposed using generative models to produce training data for text-based
applications~\cite{snorkel17}, annotation of images and videos still relies
on manual annotation by humans~\cite{papadopoulos2017extreme}. The human cost
is high. For example, 2.3 hours of urban video footage took 400 man-hours to
label.  {\em Second} and more importantly, most existing datasets, labeled or
unlabeled, come from curated sources. They have inherent biases that can
produce recognition failures when applied to raw data in the
wild~\cite{bias}.  For example, labeled images often come from edited photo
streams like Flickr, which introduce biases in lighting conditions, camera
angle and placement, or specific subjects. In contrast, models trained on
labeled training data from the same domain as classification inputs show
significantly higher accuracy and/or run-time efficiency over those trained
on curated datasets~\cite{hoffman1,kang2017noscope}.

Prior work on transfer learning addresses the first problem of data scarcity,
where a ``teacher model'' trained by a trusted
party with access to large-scale data can be shared with many general users,
who then use smaller local, targeted training data to incrementally train the final
layers of the model, producing a ``student model.'' Today, transfer learning
is recommended by most of the major deep learning frameworks, including Google Cloud ML,
Microsoft Cognitive Toolkit, and PyTorch from Facebook.

However, this does little to address the problem of training set
bias. For example, a company can customize a large pretrained facial
recognition model with images of its employees, but this student model will
produce significant recognition errors if it is trained with well-lit,
perfectly framed headshots against a white background. This issue of {\em
  domain bias} is a known problem, and there are efforts in the ML community
to address it using domain adaptation~\cite{hoffman1,hoffman2} and few shot
learning~\cite{oneshot,fewshot}.

\para{Automated Annotation during Image Capture.}
We believe that in many applications, \emph{e.g.} pedestrian and vehicle detection,
models can be extremely sensitive to domain bias, and training on labeled data
from the same precise physical context as future inputs can greatly improve
classification accuracy. This is a level of sensitivity that exceeds the
goals of existing domain transfer techniques~\cite{hoffman1,hoffman2}.  This
motivates us to explore a different approach, 
where we automatically generate annotated images using a coordinated
infrastructure of wireless receivers and image capture devices (digital
cameras). 

\begin{figure}[t]
    \centering
    \includegraphics[width=0.48\textwidth]{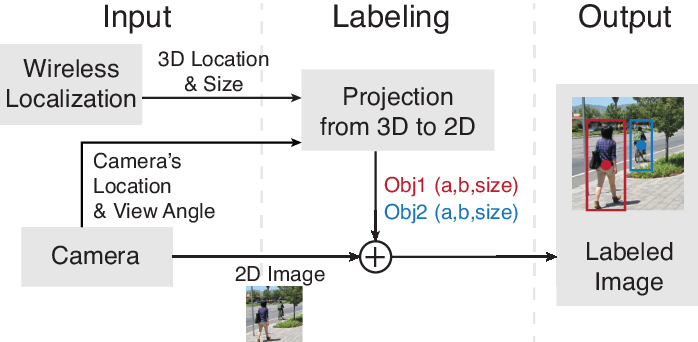}
    \caption{\name: Self-annotating image generation by combining camera with
  passive wireless localization.} 
    \label{fig:system}
\end{figure}

We believe advances in wireless localization have made it possible for (existing)
wireless infrastructure to precisely compute the 3D location of a passive
wireless device in both indoor and outdoor areas. Combined with an image
capture system, this enables the automated annotation and labeling of images
and videos as they are captured by cameras (see Figure~\ref{fig:system}).

We propose \name, a system that {\em automatically} annotates images as they
are being captured by a camera. Our insight is that locations of targets
(\emph{e.g.} pedestrians, cars, buses, drones) on a 2D captured image can be
computed from their physical 3D locations relative to the camera. New
wireless chipsets enable a feature called fine timing measurement
(FTM)~\cite{intel16ftm}\footnote{FTM is supported by Intel 8260 WiFi chip and
Android P OS.}, allowing a wireless transmitter colocated with (or
even inside) the camera to send out probes that trigger responses from nearby
WiFi devices at the hardware level. Any users or vehicles with an FTM-enabled
802.11mc device can be accurately localized, and its position on a newly
captured image can be computed using a 3D to 2D projection.

\para{Benefits.} Wireless device localization has a number of distinct
advantages over alternatives. First, this approach requires no active participation
from the participant, compared to systems where participants don wearable
RFIDs. Second, it has significantly longer range compared to RFIDs and
similar devices. Third, wireless devices responding to FTM probes will
return a device identifier, which can be used to correlate the same device or user
{\em across} different images, {\em e.g.}, temporal correlation of moving users
across images in a sequence, or geographic correlation of the same user from
images taken from different perspectives. Finally, device identifiers can be
used to support ``user-driven'' privacy policies, including the ability to
``opt-out'' of image annotation for privacy reasons. 

Unlike prior works that transfer labels from other
domains~\cite{hoffman1,hoffman2}, our goal is to automatically generate
labeled ``ground truth'' images in the same domain and setting as future
inputs. This minimizes any bias between training data and test data,  that If
successful, this would provide a low-cost mechanism for generating 
large volumes of labeled data for tasks such as pedestrian and vehicle
detection, object recognition (of connected vehicles), and facial recognition
(of wireless users).

Our work makes four key contributions. 
\begin{packed_itemize} \vspace{-0.05in}
\item We propose the use of 802.11mc FTM-enabled wireless devices, in
  coordination with digital cameras, to capture and automatically annotate
  users and vehicles with wireless devices. This system allows recognition
  systems to be trained or customized using domain-specific images with
  minimal bias.
\item Using pedestrian and vehicle detection as a case study, we identify
  practical problems in deploying automated image annotation. We identify
  four types of mismatch errors between human-labeled image data and
  automated annotations, and discuss potential efforts to address each.
\item Our empirical measurements show that pedestrian detection requires high
  quality labels beyond the precision of default 802.11 FTM hardware
  settings.  But we show via emulation that the quality of labels improves
  significantly with tuned hardware settings, indicating the feasiblity of
  our approach in the near future.
\item We recognize the seriousness of issues of participant privacy and
  consent in automated image annotation, and present some initial discussion
  in \S\ref{sec:discussion}.
\end{packed_itemize}\vspace{-0.05in}

\section{Motivation and Related Work}
\label{sec:motivation}
In this section, we perform experiments on real datasets to evaluate the
impact of physical context on image recognition accuracy. We
then discuss and differentiate our proposal from related work in the space.

\begin{figure}[t]
  \centering
    \includegraphics[width=0.25\textwidth]{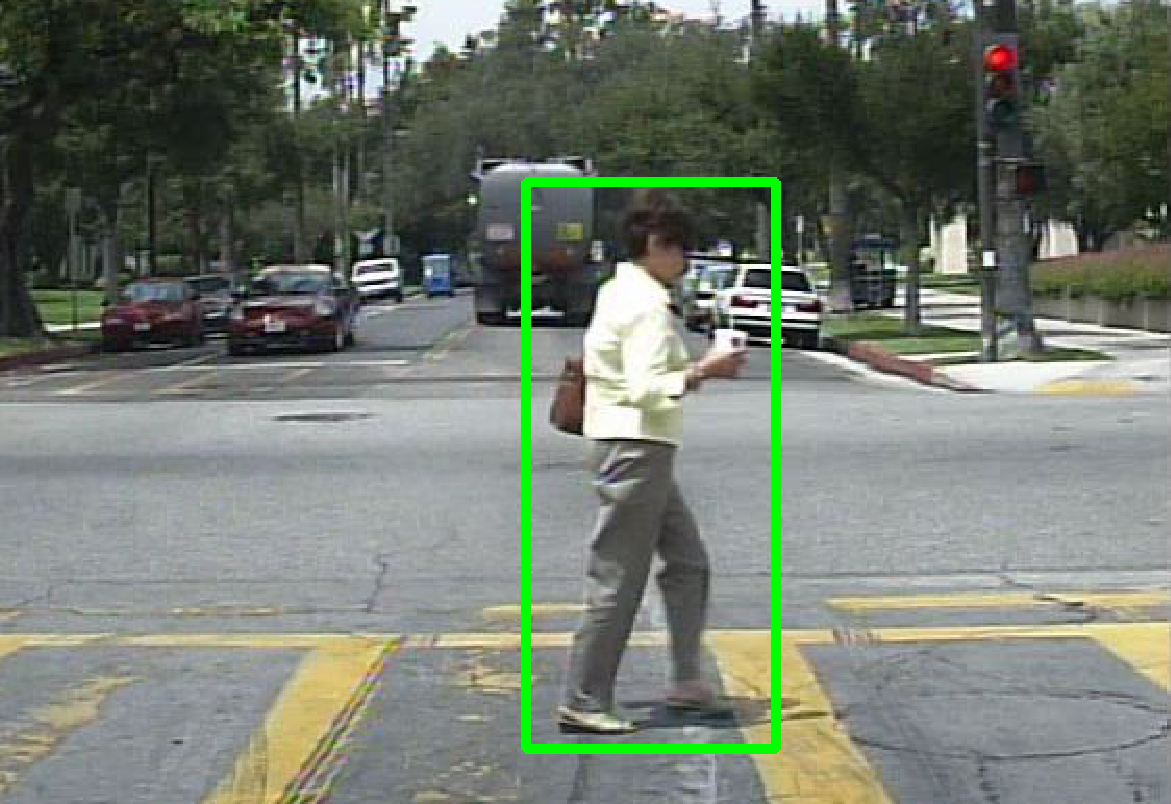}
    \caption{Manually labeling a pedestrian by a bounding box.}
    \label{fig:boundingbox}
\end{figure}

\subsection{Background: Pedestrian and Vehicle Detection}
Our experiments focus on two popular computer vision tasks: pedestrian
detection and vehicle detection.  Here we briefly introduce them and the
corresponding datasets used for ML model training and testing.  Both tasks
are critical components for modern applications like self-driving vehicles
and urban traffic management in smart cities.  For both tasks, object
annotation is an effort-intensive human task, where training data is created
by annotators who mark the boundary of each pedestrian in images using a
bounding box (Figure~\ref{fig:boundingbox}).

\para{Key Datasets.}  Table~\ref{tab:datasets} lists the details of the above
datasets.  For pedestrian detection, the most popular and annotated dataset
is the Caltech Pedestrian Dataset~\cite{dollar2012pedestrian}. It contains
2.3 hours of urban video footage collected by a single car in a day, and took
400 man-hours to label.  Another well-known dataset is the Daimler dataset,
which is collected by a car camera contributed by Daimler
Chrysler~\cite{enzweiler2008monocular}.  For car detection, there are two
popular annotated datasets for self-driving car applications:
BDD100K~\cite{bdd100k} and Udacity self-driving-car~\cite{udacity}, both are
collected by a single car moving around its local area. In addition, the
Sherbrooke dataset targets smart-city surveillance, and uses a static camera
(mounted on a street light) to capture video data.

\begin{table}[t]
\vspace{0.05in}
\resizebox{\columnwidth}{!}{
\begin{tabular}{|c|c|c|}
\hline
Dataset    & \# of images & Collected by                                      \\ \hline
Caltech    & 46,806                                                           & A moving vehicle through regular urban traffic.   \\ \hline
Dailmer    & 2,179                                                           & A moving vehicle through regular urban traffic.   \\ \hline
BDD100K    & 14,218                                                           & Crowdsourced video clips (driving).               \\ \hline 
Udacity    & 5,998                                                           & A moving vehicle during daylight conditions.      \\ \hline
Sherbrooke & 900                                                             & One fixed camera mounted meters above the ground. \\ \hline
\end{tabular}
}
\caption{Summary of evaluated datasets.}
\label{tab:datasets}
\end{table}

\para{Performance Metrics.}  Our experiment will use two common performance
metrics. The first is the average precision (defined as the mean precision at
a set of eleven equally spaced recall levels), which is widely used for
object detection tasks.  Here a higher value means more accurate
detection. The second metric is the log-average miss rate on False Positive
Per Image (FPPI) in $[10^{-2}, 10^0] $~\cite{dollar2012pedestrian}, where the
lower the value, the better the model performance.

\begin{table*}[t]
\centering

\begin{tabular}{|c|c|c|c|c|}
\hline

Dataset                                                                              & Metric            & Teacher & Non-local training & Local refinement \\ \hline
\multirow{1}{*}{Daimler}                                                             & Log avg miss rate & 37.9\%        & 10.7\%             & \textbf{4.71\%}          
                                                                                                       \\ \hline

\end{tabular}
\caption{The performance of pedestrian detection using the teacher model, the
  student model trained by task-specific but non-local data, and the
  student model further refined by local data. }
\label{table:case1a}
\end{table*}

\begin{table*}[t]
\centering

\begin{tabular}{|c|c|c|c|c|}
\hline

Dataset                                                                              & Metric            & Teacher & Non-local training & Local refinement \\ \hline
\multirow{2}{*}{\begin{tabular}[c]{@{}c@{}}Udacity\\ (moving camera)\end{tabular}}   & Log avg miss rate & 79.6\%        & 49.9\%             & \textbf{29.1\%}           \\ \cline{2-5} 
                                                                                     & Avg precision & 31.0\%        & 63.4\%             & \textbf{83.9}\%           \\ \hline
\multirow{2}{*}{\begin{tabular}[c]{@{}c@{}}Sherbrooke\\ (fixed camera)\end{tabular}} & Log avg miss rate & 60.8\%        & 88.6\%             & \textbf{12.4\%}  \\ \cline{2-5} 
                                                                                     & Avg precision & 32.3\%        & 40.5\%             & \textbf{94.5}\%           \\ \hline
\end{tabular}
\caption{The performance of vehicle detection when using the teacher model, the
  student model trained by non-local data, and the student model refined by local data.} 
\label{table:case1b}
\end{table*}

\subsection{Impact of Physical Context on Training}
Using transfer learning, we perform experiments to quantify the benefits of
training using ``local'' data, {\em i.e.\/} labeled training data with the
same physical context as classification inputs.  We compare its performance
against an uncustomized ``teacher'' model, as well as a ``student'' model
trained with ``non-local'' data, \emph{i.e.} data from a different physical context.

\para{Pedestrian Detection.}  The best performing pedestrian detection system today
is the Regional Proposal Network (RPN)~\cite{kaiming16}, a model built
on Faster
R-CNN~\cite{ren2015faster} for generic object detection.   For our
evaluation,  the RPN model trained on the generic object dataset (PASCAL
VOC~\cite{everingham2010pascal}) serves as the Teacher model.   Next,
we use a sampled portion of the Caltech dataset (40K images)
to train a student model. Since our final testing dataset is Daimler,
this student model is an example of student models trained using
task-specific but non-local data.  In the third step, we use a portion of
the Daimler dataset (800 images) to further refine the previous
model. We refer to this step as the ``local refinement.''    Finally,
we test the above three models: {\em teacher}, {\em non-local
  training}, {\em local
refinement} on the images in the Daimer dataset that were not used during local refinement. 

Table~\ref{table:case1a} shows the pedestrian detection performance (log-avg
miss rate).  When the student model is trained using task-specific but
non-local data, detection performance improves compared to the generic
teacher model. More importantly, the biggest performance improvement comes
from model refinement by training using the local dataset.

\para{Vehicle Detection.}  We perform a similar process on a vehicle
detection task. Here we use both Udacity and Sherbrooke datasets as test
datasets.  Our teacher model is the Faster R-CNN trained on the PASCAL VOC
dataset.  We then apply transfer learning on the teacher model, and use the
BDD100K dataset to create a student model (trained on task-specific but
non-local data).  Finally, we use a small portion of the Udacity dataset to
refine the above student model (local refinement), and repeat the same for
Sherbrooke.

Table~\ref{table:case1b} lists the average precision and log-avg miss rate
values. We observe the same trend as the pedestrian detection example in
Table~\ref{table:case1a}.  Here the improvement is particularly visible for
Sherbrooke, where the precision improves from 40.5\% to 94.5\%.

Again, the improvements from using ``local'' training data are extremely
large, and underscore the importance of training/customizing models using
data with the same physical context as input data.

\begin{table*}[]
\centering
\begin{tabular}{|c|c|c|c|c|}
\hline
Local refinement & None   & using model-generated labels & using RF labels & using accurate labels \\ \hline
Udacity                                                    & 63.4\% & 53.2\%                                                                  & 82.6\%          & 83.99\%               \\ \hline
\end{tabular}
\caption{Using model-generated annotations to fine-tune the student
  model actually leads to performance degradations. Values are
  average precision. }
\label{tab:case1c}
\end{table*}

\subsection{Related Work}

Our work differs from existing ML directions to reduce reliance on
labeled training data, which we summarize below. 

\para{Automatic Annotation.} Some prior works annotate objects or gestures by
physically tagging them with RFIDs~\cite{kim10_rfid} or magnetic
sensors~\cite{garcia2017first}. These require active
participation by the target and significantly limit scalability and
applicable uses.

In contrast, our goal is to scalably produce labeled images of {\em passive}
targets ({\em e.g.\/}, people, pets with collars, vehicles that {\em already} carry
WiFi devices).  Our work also differs from automatic image annotation that uses visual features ({\em e.g.\/},
color, texture, shape) to generate image labels. It requires complex generative
models that are hard to
build~\cite{zhang2012review}. 

\name\ uses target
positions reported by wireless localization to annotate them. This
differs from existing works~\cite{kang2017noscope,zeng2014deep} that run a generic detector ({\em e.g.\/}
the teacher model) on the local (unlabeled) data to produce its
annotations.  We compare \name\ to the model-generated annotation
approach using the Udacity dataset.  For latter, we use the teacher model (Faster
RCNN with ResNet) to produce model-generated annotations. 

Table~\ref{tab:case1c} lists the average precision of vehicle
detection using the Udacity dataset.  Interestingly, using
model-generated annotations to refine the student model actually leads
to degraded performance compared to the teacher model.  This result
demonstrates the importance of using real, accurate annotations. 

\para{New Model Architecture.} 

{\em Transfer learning} and {\em semi-supervised
  learning}
use local labeled data
to adapt well-trained generic models to new scenarios.  {\em Self-taught learning} and {\em
  unsupervised feature learning} learn
features from unlabeled data, but still require a sizable amount of labeled
data to train the classifier.

Finally, \emph{weakly supervised learning}~\cite{zhou2017brief}
reduces labeling complexity by using coarse-grained
 labels ({\em e.g.\/}, image-level labels without object bounding
 box), but has
 limited applicability.  Our work differs by taking a different (and
 complementary) perspective, {\em i.e.} removing labeling overhead via automation.

\section{\name: AuTomated Image Annotation}
\label{sec:background}
To enable efficient model training, \name\ integrates 
operations of a camera and a wireless networking infrastructure to automate the
process of annotating targets on images 
captured by the camera.  Specifically, \name\ reuses (existing) wireless
networking infrastructures, {\em e.g.\/} city-wide WiFi networks, to perform
passive localization on targets that carry wireless devices, and translates
the localization results into annotations on the image captured in
cohort. 

In this section, we discuss \name's basic concept and deployment cost.  We also present its benefits over human-based image
annotation,  its implications on ML application
development, and its deployment requirements and limitations.

\subsection{Overview}
\name\  leverages the fact that locations of targets
({\em e.g.\/}, pedestrians, cars, buses, bicycles, drones) on a 2D image can be derived from their physical 3D locations
w.r.t. the camera. While human locates targets on 2D images, RF localization can directly estimate each target's 3D
physical location, and then label the target on the image
by projecting its 3D location to the 2D image. If the target's physical size
is known or can be estimated,  we can use the same projection to build its bounding box on the image.

Figure~\ref{fig:system} illustrates the process of \name\ in the context of
pedestrian detection. It is very simple: the system takes as input the 3D location of each pedestrian
(\yzedit{derived by wireless localization}) and the camera image captured
at the same time. Using information
of the camera (location, view angle) and the environment (road
elevation, etc.), the system first projects each 3D location to a 2D point on the
image as the target
center. It then crafts a 3D body box based on the average human
height and body
aspect ratio,  and projects it to a 2D
bounding box based on the target's 3D location (depth).

We note that \name\ is not yet a {\em universal} solution
for annotating the entire image content.   It can
only label targets identifiable by wireless localization, {\em e.g.\/},
a person carrying a smartphone, but not an animal (unless the animal
wears a collar that contains a WiFi chip). Since many users do carry
smartphones and many vehicles are becoming connected, \name\ is particularly
applicable to vision tasks like self-driving cars and smart-city video
survelliance.  

\name\ also differs from sensor fusion, {\em e.g.\/} the previously proposed
RGB-W feature~\cite{alahi2015rgb} that combines
smartphone's wireless
signal strength data with camera images to improve detection accuracy.
Sensor fusion does not address the fundamental problem
of image annotation but leverages additional feature input to improve
detection accuracy. 

Finally, as we mentioned in \S\ref{sec:intro},  \name\ does not imply that
wireless localization will replace camera for certain vision tasks like
pedestrian and vehicle detection.  This is because wireless localization is
not ubiquitous among all the objects, while camera is ubiquitous and will
remain as the prevalent technology for many applications.

\subsection{Practicality and Cost}
\label{subsec:req}
\name\ requires tight coordination  between the camera and the localization
system. Thus one might question its practicality and deployment cost
due to the {\em extra} localization system.  We show that such
overhead can be minimized or even completely removed by reusing wireless networking
infrastructures or camera's on-board wireless radios.

\para{``Zero-cost'' Localization by Reusing Networking Infrastructure.}  Advances in mobile
computing have made precise localization of wireless devices feasible
by reusing existing infrastructure developed for networking, {\em
  e.g.\/} city-wide WiFi networks. With the recent firmware update~\cite{google11mc},
Google Android devices (with Android P) already support a WiFi localization
protocol (802.11mc) that achieves meter-level accuracy in both indoor
and outdoor environments.  In the near 
future, cellular providers will deploy 5G networks with mmWave radios,
enabling localization at a cm-level accuracy.  Therefore, by leveraging existing and upcoming wireless networking
infrastructure, the cost of localization in \name\ could reduce to zero.

\para{Direct Deployment on Camera's On-board Radios.}   Cameras for self-driving
cars and smart-city applications are now equipped with GPUs and
wireless networking systems (WiFi and Cellular).  This means that \name\ can be implemented directly on cameras without requiring
external  infrastructure.

\subsection{Benefits}
By combining image capture and annotation, \name\ offers five unique
advantages over human annotation. 

\para{Volume.}  As data labeling is fully automated, the size, number
and diversity of training data
will no longer be constrained by human
  labor and can become arbitrarily large.

\para{Real-time annotation while collecting data.} In \name\, annotation
works in unison with image capturing and thus the two tasks run
simultaneously. Thus the annotated images can be immediately fedback to the
machine learning model for online refinement. 

\para{Adding location/depth to Images.}  Each annotation produced by \name\  includes
the 3D physical location of the target,  thus adding depth to 2D images
(using normal cameras). Human
annotation cannot do so.

\para{Temporal Tracking.}  With consent, \name\ can
  track  each identified target over time, producing fine-grained, context-rich
  labels on the target.  For example,  one can infer the moving
  speed and context of a target from its sequence of location data ({\em e.g.\/},
  standing, walking, running, biking), and use them to produce fine-grained
  annotations on the target. 

\para{Cross-perspective Correlation.} \name\ can extract {\em
    hidden} identity of the targets that are hard to identify from the camera
  data. For example, it can use captured WiFi MAC addresses as identity
trackers (assume no MAC randomization). After recognizing the same identity across camera and time, one can correlate
these images together to build a comprehensive view of the target.

\subsection{Implications on ML Applications}

Aside from boosting the number and size of
labeled training data,  \name\ can also facilitate development of
highly complex ML applications. Below are some examples.

\para{3D Face Models.}   Computer vision tasks like
person re-identification
 and multi-view face detection and recognition face
significant challenges since they require training
data on each target's identity with images across frames,
cameras, and locations.  The labeling task is extremely
difficult for human annotators since viewpoint, background, lighting and illumination can
change significantly across images.  With \name, we can track targets by their
device identities,  and  {\em automatically} build a comprehensive view
of the target across many images.  This helps to create a
large database of
3D, multi-view facial data for individual users.

\para{Human Action Recognition.} A conventional method for building
training data for human activity recognition is to ask volunteers to
perform predefined actions in front of
cameras~\cite{schuldt2004recognizing}, which clearly does not
scale. \name\ can automatically identify and
label activities based on each target's location data over time. For
example,  our initial experiments find it can separate bikers and
runners from stationary users
and walkers based on their moving speeds.  It can also use target
location ({\em e.g.\/}, bike
lane vs. side walk) to separate bikers and runners who move at similar
speeds.

\para{Abnormal Event Detection in  Video Surveillance.} The
physical location and trajectory can also be used to
identify and label abnormal events for video
surveillance. For example, one can create
detailed labels when users stand or run in the middle of the
street, or follow an unusual route.

\para{Scene Recognition.} \name\ can label physical
objects, from vehicles (cars, buses and trucks), robots and drones that
are equipped with RF devices, to doors with WiFi smart video
doorbells.
The same benefits of adding depth, temporal tracking and
across-camera integration can be utilized to recognize
and label scenes on images.

\subsection{Requirements and Limitations}
Deploying \name\ in practice imposes
three key requirements on the underlying localization system.
{\em First}, to achieve sufficient coverage,  the localization system
should be passive, not requiring targets to
actively communicate or synchronize with the system.  {\em Second},  it also  needs to
support a {\em range} similar to that of camera ($\approx$ 60m for outdoor clear
view~\cite{dollar2012pedestrian}).  It needs to be time-synchronized with the camera (at the level of camera frame
rate). {\em Third}, it needs to offer high precision.

\name\ also faces two key limitations when compared to human annotation of
images.  The first is inherent, and the second depends on
the performance of the underlying target localization technology. 

\begin{packed_itemize} \vspace{-0.06in}
\item \name\ cannot annotate existing camera datasets (that do not contain
  localization data);

\item \name\  cannot recognize and label targets that do not have
wireless devices, and errors in localization results will translate into
erroreous data labels. We further discuss and address this limitation in the
following sections.\vspace{-0.06in}

\end{packed_itemize}


\begin{figure}[t]
  \centering
    \includegraphics[width=0.45\textwidth]{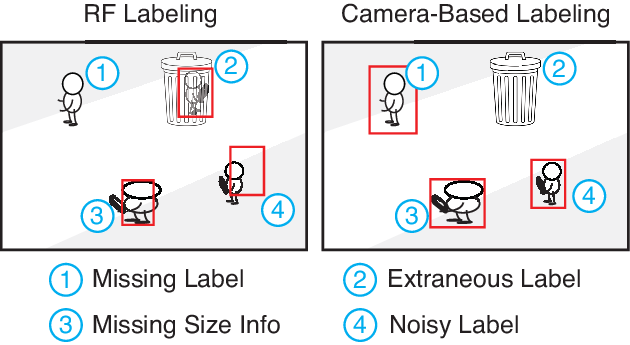}
    \caption{Four types of mismatch between objects recognized by camera and
      those recognized by \name's wireless localization.}
    \label{fig:artifact_rf}
\end{figure}

\section{The Problem of Label Mismatch}
\label{sec:challenge}



The key performance challenge facing \name\ is the  potential
mismatch between targets captured by camera and that detected by \name's
wireless localization.  
 We categorize such mismatch into four types
(illustrated in Figure~\ref{fig:artifact_rf}).

\para{Type \raisebox{.5pt}{\textcircled{\raisebox{-.9pt} {1}}}: 
Missing Labels due to Partial Wireless Coverage.}  As mentioned earlier, not
every target can be detected and localized by wireless localization, {\em
  e.g.\/}, outdoor localization systems can only localize
targets who carry the required RF device and are in range.
Therefore, \name\ could miss some targets. 

We expect that these missing labels will have minimal impact on the ML model 
performance because they can be compensated by annotating more images (easy to achieve by \name).  Of course, this assumes
\name's wireless localization system either does not impose any {\em bias} on targets, or any
such bias can be addressed by the DNN model itself, and that training cost does not grow drastically with the number of images. 


Using the pedestrian and vehicle detection examples in
\S\ref{sec:motivation}, we study the impact of 
missing labels and test our hypothesis.  For each dataset, given a wireless coverage ratio of $p$
(\%), we create \name\ annotations by randomly
sampling the ``ground truth'' annotations in the dataset.  

\begin{table}[h]
\centering
\begin{tabular}{|c|c|c|}
\hline
\multirow{2}{*}{Dataset} & \multicolumn{2}{c|}{Coverage $p$} \\ \cline{2-3} 
                         & 30\%          & 100\%         \\ \hline
Caltech                  & 22.2\%        & 17.3\%        \\ \hline
Dailmer                  & 8.60\%        & 4.71\%        \\ \hline
Udacity                  & 31.9\%        & 29.1\%        \\ \hline
\end{tabular}
\caption{Detection performance for $p = 30\%$ and $100\%$.}
\label{table:partial_coverage}
\end{table}

Table~\ref{table:partial_coverage} lists the detection performance (log avg miss
rate) for different $p$ values.  For each dataset, we keep the number
of training images constant.  We see that the impact of missing labels
is small for Udacity, but more visible for Daimler and Caltech.

\begin{figure}[t]
\centering 
 \includegraphics[width=0.4\textwidth]{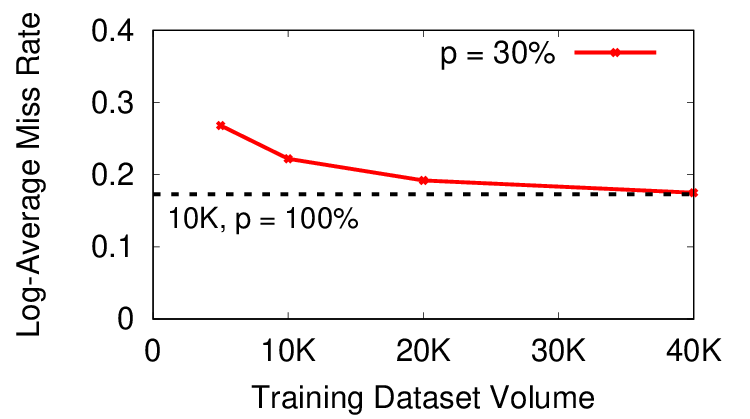}
    \caption{Partial RF coverage $p$ affects detection 
     performance, 
     but gets compensated by annotating more images. The results are
     for the Caltech dataset.}
    \label{fig:partialpen_more}
\end{figure}

Next, for the Caltech dataset, we study the miss rate of $p$=30\% but vary the  number of training
images.  Figure~\ref{fig:partialpen_more} reconfirms that for
a given number of training images, {\em e.g.\/} 10k,  lower $p$ leads to less
accurate pedestrian detection. But such degradation can be compensated by
adding more annotated images: with 10k more
training images (5 minutes of video data), the miss rate reduces to 18\%.

We also study the impact of potential {\em bias} imposed by \name. For pedestrian detection,  \name\ cannot label children
since they do not normally carry wireless devices.  We verified that today's
pedestrian detection model~\cite{kaiming16} can detect children
accurately even when all labeled pedestrians are adults. 
This is because the DNN model treats children as a scaled down version of
adults.

\para{Type \raisebox{.5pt}{\textcircled{\raisebox{-.9pt} {2}}}: 
Extraneous Labels due to Camera Occlusion.} 
Since wireless signals often can penetrate or go ``around'' obstacles,
\name\ may locate and label targets behind obstacles. Yet cameras can
only capture the obstacles or parts of the target, {\em i.e.\/}, {\em camera
  occlusion}. Since both full and partial occlusions should not be used during model
training~\cite{dollar2012pedestrian},  \name's annotation of occluded 
targets need to be identified and removed. 

\begin{figure}[t]
  \centering
    \includegraphics[width=0.48\textwidth]{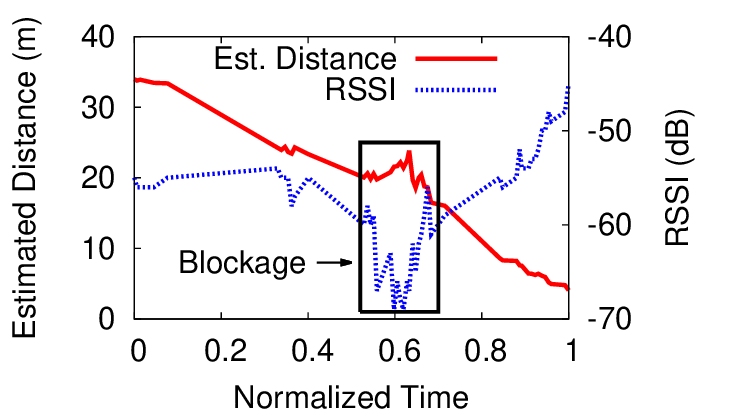}

     \caption{Wireless localization can potentially identify when a target gets blocked by
       examining the signal strength value and the localization result over time.}
     \label{fig:infer_blockage}
 \end{figure}

To detect camera occlusion, one potential direction is to analyze captured wireless signals and
localization results over time.  Intuitively,  obstacle blockage leads to large degradation in
signal strength, and the use of NLoS paths will produce much longer range 
estimation. Figure~\ref{fig:infer_blockage} shows an example where a
pedestrian was blocked by an obstacle for a period of time.  These 
artifacts will appear as
``anomalies'' in the time sequence of signal measurements and
localization results, which can be used to detect occlusion.

\para{Type \raisebox{.5pt}{\textcircled{\raisebox{-.9pt} {3}}}: 
Missing Size Information.}  While a human annotator will draw a
bounding box around each detected target, \name's wireless localization does not offer such
information. For pedestrian detection, this
can be overcome by first establishing 
an estimate of the target size ({\em e.g.\/}, the average human
height \zjedit{1.76m~\cite{fryar2016anthropometric}} and the average aspect
ratio \zjedit{0.41}~\cite{dollar2012pedestrian}), 
 and then projecting the physical bounding box to the
camera bounding box based on the target's relative location to the
camera. This is a reasonable estimate,
 because today's pedestrian detection models~\cite{kaiming16} also resize the 
 bounding boxes based on the same average aspect ratio (0.41). 

A related challenge  is how to distinguish between 
targets that carry the same type of wireless devices, {\em e.g.\/}, human with WiFi
vs. machines/vehicles with WiFi.  One can analyze the MAC
address\footnote{Existing works have shown that one can infer the device type
  from the Organizational Unique Identifier (OUI) field of the MAC
  address~\cite{privacy14iccst}.} or the physical
location and trajectory data obtained via localization. For example, human
users will most likely travel on sidewalks while vehicles stay in their
lanes. The two will also have different movement patterns.

\para{Type \raisebox{.5pt}{\textcircled{\raisebox{-.9pt} {4}}}: 
Noisy Labels due to Localization Errors.}  In this case, a target is captured
by both \name's wireless
localization system and the camera.  But its annotation derived by \name\
deviates from the ground truth label\footnote{Here we assume
  that human labeling is always accurate, which is in general not true in
  reality.} due to localization errors 
(and bounding box estimation errors).  In this case, localization errors include both {\em angular}
and {\em depth} errors. The angular error ``shifts'' the location of the 
annotation on the image, while the depth error ``modifies''  the bounding box size.

\section{Empirical Analysis of Noisy Labels}
\label{sec:noisylabel}
While the first three types of mismatch discussed in \S\ref{sec:challenge} are results of inherent difference
between camera and wireless localization,  the noisy labels are caused by
localization errors of \name's underlying wireless localization system.  In
this section,  we empirically analyze the impact of
noisy labels using a specific wireless localization system.

 \begin{figure*}[t]
 \centering
  \begin{minipage}{0.3\textwidth}
    \centering
   \vspace{0.05in}
    \includegraphics[width=1\textwidth]{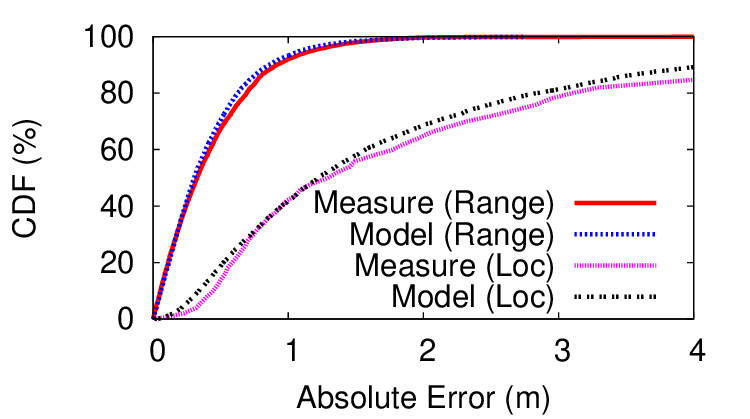}
    \caption{Our
    error model aligns with the testbed measurements.}
    \vspace{-0.1in}
    \label{fig:errormodel}
  \end{minipage}
  \hfill
  \begin{minipage}{0.3\textwidth}
    \centering
    \resizebox{1\columnwidth}{!}{%
    \begin{tabular}{|c|c|c|cc|}
    \hline
    \multirow{2}{*}{} & \multirow{2}{*}{\begin{tabular}[c]{@{}c@{}}\# of\\
        TXs\end{tabular}} & \multirow{2}{*}{\begin{tabular}[c]{@{}c@{}}sample\\
        rate\end{tabular}} & \multicolumn{2}{c|}{localization error (cm)} \\ \cline{4-5}
     &  &  & \multicolumn{1}{c|}{\ median\ } & 95\% \\ \hline
    S0 & 2 & 256 & \multicolumn{1}{c|}{132.0} & 462.8 \\ \hline
    S1 & 4 & 2048 & \multicolumn{1}{c|}{31.8} & 93.8 \\ \hline
    S2 & 6 & 2048 & \multicolumn{1}{c|}{24.6} & 63.8 \\ \hline
    S3 & 6 & 5012 & \multicolumn{1}{c|}{16.2} & 42.0 \\ \hline
    \end{tabular}
    }
    \caption{\yzedit{802.11 FTM} localization under different
      configurations. S0 is our testbed, S1--S3 are projections.} 
    \label{table:errormodel}    
  \end{minipage}
  \hfill
  \begin{minipage}{0.33\textwidth}
    \centering
    \includegraphics[width=1\textwidth]{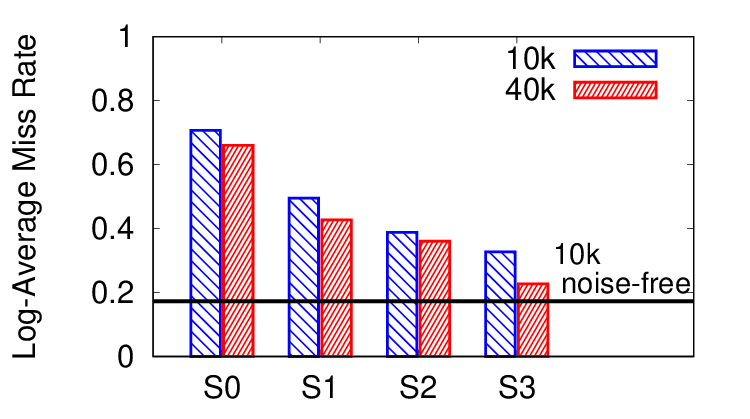}
    \caption{Impact of noisy labels on pedestrian detection w/ 10k and 40k training images.}
    \label{fig:impactped}
  \end{minipage}
\end{figure*}

\subsection{Passive Localization via WiFi  FTM}
\label{subsec:11mc} 
\name\ employs a recent development of passive WiFi localization:  IEEE 802.11mc with 
fine timing measurements (FTM)~\cite{802.11mc,intel16ftm}, 
or \yzedit{802.11 FTM} 
 in short.  
  802.11 FTM uses time-of-flight (ToF) for ranging and trilateration for
  localization. Each camera (or a co-located 802.11 FTM access point) will
  broadcast FTM beacons continuously.  A target that supports FTM, when in
  range, will automatically respond to these beacons. The camera then
  estimates the distance to the target by measuring the round-trip
  time (RTT). 
    Differed from existing works on ToF ({\em e.g.\/},~\cite{fadel15nsdi, yunfei17mobicom,
    deepak16nsdi,xiong15tonetrack}), \yzedit{802.11 FTM} operates on simple sine waves. 
   Thus its accuracy depends on the precision of hardware timing 
    and RTT estimation.

  It should be noted that 802.11 FTM  is a {\em hardware} feature and does
  not require the target to set up network-level connectivity or
  synchronization\footnote{It estimates RTT locally in an asynchronous
    manner.}  with the camera (or access point).  This feature enables
  physical layer timing with a picosecond-level accuracy,  and thus offers
  sub-meter-level ranging accouracy.  Another important factor is that its
  ranging/localization accuracy is {\em location-independent}, as long as the
  signal can reach the target.  Finally, 802.11 FTM is already supported by off-the-shelf WiFi 
  chipsets, {\em e.g.\/} Intel 8260 and 
  Google Android P~\cite{google11mc}. Today, the Intel 8260
    chipset costs less than \$20, and supports a range of at least $100m$~\cite{11mc_range}.

\para{Modeling 802.11 FTM Localization Errors.}  Using detailed testbed
measurements, we build an empirical model on the localization error of 802.11
FTM.  Our measurements used  three Dell XPS 13 laptops with the Intel 8260 chipset (2 as transmitters
 and 1 as target) placed on typical streets of 
 approximately $5\times40$ m$^2$ in
 size.  After analyzing 10,000 testbed measurements with 
 varying target locations, we found that the ranging error follows the (folded) {\em t location-scale distribution}~\cite{tlocationscale} with
 zero mean and $0.54m$ standard deviation, and the localization error follows
 the gamma distribution.  Both distributions remain invariant
 (with 91\% confidence) despite changes in weather, target orientation
 and location, distance to transmitters, and
 hardware. We also confirmed that the error models match the testbed data in
 Figure~\ref{fig:errormodel}. 

The above model assumes 2 transmitters, a sampling rate of 256 beacons per
 localization instance  
 \yzedit{(a hardcoded limit for our testbed hardware)}.
 We also simulated more sophisticated hardware
 configurations by increasing the number of transmitters and the sampling
 rate
  and found that they do not change the model
 distributions, only the model parameters. Table~\ref{table:errormodel} summarizes the model parameters for four
different configurations: S0 is our current testbed, S1-S3 are projected
values of more sophisticated hardware configurations.  For the S3
configuration, the median localization error reduces to $16.2cm$.

\subsection{Impact of Noisy Labels}

Given the above localization error model, we can now evaluate \name\ using existing
pedestrian detection datasets like Caltech, producing noisy labels of
pedestrians on the images.  To do so, we first estimate the 3D physical
location of each ``ground-truth'' label in the Caltech dataset (we 
approximate this value using the pinhole camera
model~\cite{pinhole} assuming a pedestrian height of
1.76m~\cite{fryar2016anthropometric}, and then add 10\% random variation). We
then inject our model-generated localization error on each instance and
project it back to the same 2D image to create noisy labels produced by
\name.

\label{subsec:noiseimpact}
We use these noisy labels to train a pedestrian detection model, and test on
other segments of the Caltech dataset. 
Figure~\ref{fig:impactped} compares the log-average miss rate of the
pedestrian detection 
model trained with 5 minutes and 20 minutes of video images, for different
\name\ localization configurations (S0-S3).  Our baseline is the model
trained with 5 minutes of video images with error-free annotations.

We make two key observations. {\em First}, for pedestrian detection, the current DNN
model is sensitive to noisy labels produced by wireless localization.  Under our
basic FTM configuration (S0, 1.32m median localization error), the miss rate rises to more
than 70\% compared to 17.3\% under noise-free labels.  It drops back to 20\%
after improving hardware and infrastructure density (S3, 16.2cm median 
error) and adding more labeled data.  This result confirms the initial
feasibility of \name\ on practical ML tasks. 

{\em Second}, we found that
angular error is the dominant factor for 
performance degradation (compared to depth error).  That is, placing a bounding box at the
wrong 2D location on the image leads to much higher damage than wrongly sizing the
bounding box.  This observation could be further utilized to reduce the
impact of noisy labels. 

\para{Experimenting with Other Datasets.}  We focus our evaluation on
the Caltech dataset because it includes the detailed information about
the camera angle and position so we can accurately inject localization
errors into the dataset.  The other datasets like Daimler for
pedestrian detection and Udacity/BDD100K for car detection do not
contain such camera information and thus cannot be used in our current
evaluation. As ongoing work, we are working on locating these
configurations and identifying other feasible datasets, and use them
to perform a more comprehensive set of evaluation.

\section{Discussion}

\subsection{Reducing the Impact of Noisy Labels}
\label{sec:reducenoise}
Our results show that the most immediate challenge facing \name\ is
noisy labels. We now discuss three orthogonal and complementary directions to
address this problem. Further research efforts are needed in these
areas.

\para{Advancing Outdoor Localization.} For pedestrian detection, \name\ 
requires precise outdoor  localization (tens of $cm$) at a $60m$
range. Today's solutions were never designed with this level of accuracy in
mind. The straightforward solution to advancing outdoor localization is to
motivate industry to increase AP density and upgrade RF hardware, {\em
  e.g.\/}, increasing 802.11 FTM beacon rate from 256 to 5012 per unit (S3 in
Table~\ref{table:errormodel}), or switching to directional mmWave radios for
localization. We can emulate some of these upgrades today,
{\em e.g.\/} using multiple 802.11 
FTM chipsets to emulate higher beacon rates; or adapt localization
methods to focus on minimizing angular errors. 

\para{Filtering Out Noisy Labels.} 
Our second approach is using data analysis to
identify ``bad'' localization instances, and ignore the corresponding
labels. As a result, each \name\ annotated image will miss some labels, which can
be compensated by labeling more images (see 
\S\ref{sec:challenge}). 

There are two potential methods for identifying bad localization
instances, depending on the data used for analysis. 
 The {\em first} directly identifies bad instances by
looking at raw localization data. Recent work achieves this by applying
unsupervised feature clustering on large-scale WiFi and cellular RSS
localization 
datasets~\cite{www2017}. The
system can effectively identify and remove bad localization instances. It would be interesting to study whether
the same approach can be used on FSM localization data.

The second and complementary approach is to cross-validate each RF label using its
corresponding visual content, {\em e.g.\/}, the image content inside the
bounding box. Intuitively, an accurate label will create a bounding box
around an object, which ``stands out'' from the surrounding
background.  In computer vision, this is captured by a metric called {\em
  objectness score}, which measures how likely a bounding box contains an
object~\cite{alexe2010object,cheng2014bing,
ren2015faster,edgebox14}.  This method, however, cannot differentiate between
types of  objects ({\em e.g.\/} pedestrians vs. trash bins). One can partially
compensate by considering a sequence of images and leveraging temporal
correlation of pedestrian movement. For example, recent work has leveraged
view synthesis~\cite{zhou2017unsupervised} to estimate depth and
motion of targets, which can be combined with localization results for
cross-validation.

\para{Error-Resilient DNN Models.} Our third approach is to apply architectural modification
to existing models so that they can
tolerate noisy labels. This is a well-studied topic in the ML 
community, 
especially for classification
tasks.

\name\ brings new opportunities in this domain, since wireless 
localization can simultaneously provide multiple forms of labels of different
accuracy levels, {\em
  i.e.\/} minimum number of pedestrians in the image (most accurate), depth of each pedestrian, and
3D physical location of each pedestrian (least accurate). Our RF data
analysis can also offer confidence scores for each label~\cite{snorkel17}. How to build robust
models for these new scenarios is an interesting open research question.

\subsection{Privacy Opt-out via Device-based Localization}
For passive annotated imaging to move forward, participant consent and
privacy is a critical issue that must be addressed comprehensively. 
We believe that \name\ can utilize device-based localization to help address such privacy
concerns.  Since only targets carrying a specific wireless device will be
recognized and labeled by the system, a user can specify her privacy
constraints to the annotation system based on her device identity, {\em
  e.g.\/}, the MAC address (using 802.11mc probing). In particular, a user can opt out
completely or at specific locations and time periods (since \name\ knows
the exact user location and time).  Such privacy protection cannot be implemented using
manual labeling.  Finally, while this new feature offers an initial start on
user privacy protection, we still need significant research efforts to
address the issue of participant privacy and consent.

\section{Conclusion}
\label{sec:discussion}

We believe advances in localization will make it possible to
precisely compute the location of a passive wireless device, thus enabling
automated annotation of some targets on images (and other datasets).  Using
case studies on pedestrian and vehicle detection, we demonstrate the
feasibility, benefits, and challenges of such concept. Our work calls for new
technical developments on passive localization, mobile data analytics, and
error-resilient ML models, as well as privacy protection during ML training.
Compared with ongoing efforts in the ML community, this approach tackles the
hard challenge of (training) data annotation from a different (and complementary)
perspective, {\em i.e.\/} removing annotation overhead via automation.

\bibliographystyle{sysml2019}

\begin{thebibliography}{45}
\providecommand{\natexlab}[1]{#1}
\providecommand{\url}[1]{\texttt{#1}}
\expandafter\ifx\csname urlstyle\endcsname\relax
  \providecommand{\doi}[1]{doi: #1}\else
  \providecommand{\doi}{doi: \begingroup \urlstyle{rm}\Url}\fi

\bibitem[goo()]{google11mc}
Previewing android p.
\newblock
  \url{https://android-developers.googleblog.com/2018/03/previewing-android-p.html}.

\bibitem[tlo()]{tlocationscale}
Statistics: t location-scale distribution.
\newblock
  \url{https://www.mathworks.com/help/stats/t-location-scale-distribution.html}.

\bibitem[802(2016)]{802.11mc}
Part 11: Wireless lan medium access control ({MAC}) and physical layer ({PHY})
  specifications.
\newblock \emph{{IEEE P802.11-REVmc}}, 2016.

\bibitem[uda(2016)]{udacity}
Udacity self-driving car dataset.
\newblock \url{https://github.com/udacity/self-driving-car}, 2016.

\bibitem[Adib et~al.(2015)Adib, Kabelac, and Katabi]{fadel15nsdi}
Adib, F., Kabelac, Z., and Katabi, D.
\newblock Multi-person localization via {RF} body reflections.
\newblock In \emph{Proc. of NSDI}, 2015.

\bibitem[Alahi et~al.(2015)Alahi, Haque, and Fei-Fei]{alahi2015rgb}
Alahi, A., Haque, A., and Fei-Fei, L.
\newblock {RGB-W}: When vision meets wireless.
\newblock In \emph{Proc. of ICCV}, 2015.

\bibitem[Alexe et~al.(2010)Alexe, Deselaers, and Ferrari]{alexe2010object}
Alexe, B., Deselaers, T., and Ferrari, V.
\newblock What is an object?
\newblock In \emph{Proc. of CVPR}, 2010.

\bibitem[Andriluka et~al.(2018)Andriluka, Iqbal, Milan, Insafutdinov,
  Pishchulin, Gall, and Schiele]{andriluka2018posetrack}
Andriluka, M., Iqbal, U., Milan, A., Insafutdinov, E., Pishchulin, L., Gall,
  J., and Schiele, B.
\newblock Posetrack: A benchmark for human pose estimation and tracking.
\newblock In \emph{Proc. of CVPR}, 2018.

\bibitem[Au(2016)]{11mc_range}
Au, E.
\newblock {The Latest Progress on IEEE 802.11mc and IEEE 802.11ai Standards}.
\newblock \emph{IEEE Vehicular Technology Magazine}, 11\penalty0 (3), 2016.

\bibitem[Banin et~al.(2016)Banin, Schatzberg, and Amizur]{intel16ftm}
Banin, L., Schatzberg, U., and Amizur, Y.
\newblock {WiFi FTM} and map information fusion for accurate positioning.
\newblock In \emph{Proc. of IPIN}, 2016.

\bibitem[Cheng et~al.(2014)Cheng, Zhang, Lin, and Torr]{cheng2014bing}
Cheng, M., Zhang, Z., Lin, W., and Torr, P.
\newblock Bing: Binarized normed gradients for objectness estimation at 300fps.
\newblock In \emph{Proc. of CVPR}, 2014.

\bibitem[Dollar et~al.(2012)Dollar, Wojek, Schiele, and
  Perona]{dollar2012pedestrian}
Dollar, P., Wojek, C., Schiele, B., and Perona, P.
\newblock Pedestrian detection: An evaluation of the state of the art.
\newblock \emph{IEEE Transactions on Pattern Analysis and Machine
  Intelligence}, 34\penalty0 (4), 2012.

\bibitem[Enzweiler \& Gavrila(2008)Enzweiler and
  Gavrila]{enzweiler2008monocular}
Enzweiler, M. and Gavrila, D.~M.
\newblock Monocular pedestrian detection: Survey and experiments.
\newblock \emph{IEEE Transactions on Pattern Analysis \& Machine Intelligence},
  12, 2008.

\bibitem[Everingham et~al.(2010)Everingham, Van~Gool, Williams, Winn, and
  Zisserman]{everingham2010pascal}
Everingham, M., Van~Gool, L., Williams, C.~K., Winn, J., and Zisserman, A.
\newblock {The Pascal visual object classes (voc) challenge}.
\newblock \emph{International journal of computer vision}, 88, 2010.

\bibitem[Fryar et~al.(2016)Fryar, Gu, Ogden, and
  Flegal]{fryar2016anthropometric}
Fryar, C.~D., Gu, Q., Ogden, C.~L., and Flegal, K.~M.
\newblock Anthropometric reference data for children and adults: United states,
  2011-2014.
\newblock \emph{Vital and Health Statistics Series}, 2016.

\bibitem[Garcia-Hernando et~al.(2018)Garcia-Hernando, Yuan, Baek, and
  Kim]{garcia2017first}
Garcia-Hernando, G., Yuan, S., Baek, S., and Kim, T.-K.
\newblock First-person hand action benchmark with {RGB-D} videos and {3D} hand
  pose annotations.
\newblock In \emph{Proc. of CVPR}, 2018.

\bibitem[Gebru et~al.(2017)Gebru, Hoffman, and Fei-Fei]{hoffman1}
Gebru, T., Hoffman, J., and Fei-Fei, L.
\newblock Fine-grained recognition in the wild: A multi-task domain adaptation
  approach.
\newblock In \emph{Proc. of ICCV}, 2017.

\bibitem[Hartley \& Zisserman(2004)Hartley and Zisserman]{pinhole}
Hartley, R.~I. and Zisserman, A.
\newblock \emph{Multiple View Geometry in Computer Vision}.
\newblock Cambridge University Press, 2004.

\bibitem[Kang et~al.(2017)Kang, Emmons, Abuzaid, Bailis, and
  Zaharia]{kang2017noscope}
Kang, D., Emmons, J., Abuzaid, F., Bailis, P., and Zaharia, M.
\newblock Noscope: optimizing neural network queries over video at scale.
\newblock \emph{Proceedings of the VLDB Endowment}, 10, 2017.

\bibitem[Kemelmacher-Shlizerman et~al.(2016)Kemelmacher-Shlizerman, Seitz,
  Miller, and Brossard]{kemelmacher2016megaface}
Kemelmacher-Shlizerman, I., Seitz, S.~M., Miller, D., and Brossard, E.
\newblock The megaface benchmark: 1 million faces for recognition at scale.
\newblock In \emph{Proc. of CVPR}, 2016.

\bibitem[Kim \& Chang(2010)Kim and Chang]{kim10_rfid}
Kim, H. and Chang, S.
\newblock {RFID assisted image annotation system for a portable digital
  camera}.
\newblock In \emph{Proc. of ICCAS}, 2010.

\bibitem[Kim et~al.(2017)Kim, Keane, Wang, Tang, Riggle, Shakhnarovich,
  Brentari, and Livescu]{kim2017lexicon}
Kim, T., Keane, J., Wang, W., Tang, H., Riggle, J., Shakhnarovich, G.,
  Brentari, D., and Livescu, K.
\newblock Lexicon-free fingerspelling recognition from video: Data, models, and
  signer adaptation.
\newblock \emph{Computer Speech \& Language}, 46, 2017.

\bibitem[Li et~al.(2017)Li, Nika, Zhang, Zhu, Yao, Zhao, and Zheng]{www2017}
Li, Z., Nika, A., Zhang, X., Zhu, Y., Yao, Y., Zhao, B.~Y., and Zheng, H.
\newblock Identifying value in crowdsourced wireless signal measurements.
\newblock In \emph{Proc. of WWW}, 2017.

\bibitem[Luo et~al.(2017)Luo, Zou, Hoffman, and Fei-Fei]{hoffman2}
Luo, Z., Zou, Y., Hoffman, J., and Fei-Fei, L.
\newblock Label efficient learning of transferable representations across
  domains and tasks.
\newblock In \emph{Proc. of NIPS}, 2017.

\bibitem[Ma et~al.(2017)Ma, Selby, and Adib]{yunfei17mobicom}
Ma, Y., Selby, N., and Adib, F.
\newblock Minding the billions: Ultra-wideband localization for deployed {RFID}
  tags.
\newblock In \emph{Proc. of MobiCom}, 2017.

\bibitem[Mallapuram et~al.(2017)Mallapuram, Ngwum, Yuan, Lu, and
  Yu]{mallapuram2017smart}
Mallapuram, S., Ngwum, N., Yuan, F., Lu, C., and Yu, W.
\newblock Smart city: The state of the art, datasets, and evaluation platforms.
\newblock In \emph{Proc. of ICIS}, 2017.

\bibitem[Papadopoulos et~al.(2017)Papadopoulos, Uijlings, Keller, and
  Ferrari]{papadopoulos2017extreme}
Papadopoulos, D.~P., Uijlings, J.~R., Keller, F., and Ferrari, V.
\newblock Extreme clicking for efficient object annotation.
\newblock In \emph{Proc. of ICCV}, 2017.

\bibitem[Ratner et~al.(2017)Ratner, Bach, Ehrenberg, Fries, Wu, and
  R{\'{e}}]{snorkel17}
Ratner, A., Bach, S.~H., Ehrenberg, H.~R., Fries, J.~A., Wu, S., and R{\'{e}},
  C.
\newblock Snorkel: Rapid training data creation with weak supervision.
\newblock \emph{PVLDB}, 11\penalty0 (3), 2017.

\bibitem[Ravi \& Larochelle(2017)Ravi and Larochelle]{fewshot}
Ravi, S. and Larochelle, H.
\newblock Optimization as a model for few-shot learning.
\newblock In \emph{Proc. of ICLR}, Toulon, France, April 2017.

\bibitem[Ren et~al.(2015)Ren, He, Girshick, and Sun]{ren2015faster}
Ren, S., He, K., Girshick, R., and Sun, J.
\newblock Faster r-cnn: Towards real-time object detection with region proposal
  networks.
\newblock In \emph{Proc. of NIPS}, 2015.

\bibitem[Sanchez et~al.(2014)Sanchez, Satta, Fovino, Baldini, Steri, Shaw, and
  Ciardulli]{privacy14iccst}
Sanchez, I., Satta, R., Fovino, I.~N., Baldini, G., Steri, G., Shaw, D., and
  Ciardulli, A.
\newblock Privacy leakages in smart home wireless technologies.
\newblock In \emph{Proc. of ICCST}, 2014.

\bibitem[Schuldt et~al.(2004)Schuldt, Laptev, and
  Caputo]{schuldt2004recognizing}
Schuldt, C., Laptev, I., and Caputo, B.
\newblock {Recognizing human actions: a local SVM approach}.
\newblock In \emph{Proc. of ICPR}, 2004.

\bibitem[Torralba \& Efros(2011)Torralba and Efros]{bias}
Torralba, A. and Efros, A.~A.
\newblock Unbiased look at dataset bias.
\newblock In \emph{Proc. of CVPR}, 2011.

\bibitem[Vasisht et~al.(2016)Vasisht, Kumar, and Katabi]{deepak16nsdi}
Vasisht, D., Kumar, S., and Katabi, D.
\newblock Decimeter-level localization with a single {WiFi} access point.
\newblock In \emph{Proc. of NSDI}, 2016.

\bibitem[Vinyals et~al.(2016)Vinyals, Blundell, Lillicrap, Kavukcuoglu, and
  Wierstra]{oneshot}
Vinyals, O., Blundell, C., Lillicrap, T., Kavukcuoglu, K., and Wierstra, D.
\newblock Matching networks for one shot learning.
\newblock In \emph{Proc. of NIPS}, Barcelona, Spain, December 2016.

\bibitem[Wang et~al.(2018)Wang, Shen, Guo, Cheng, and
  Borji]{wang2018revisiting}
Wang, W., Shen, J., Guo, F., Cheng, M.-M., and Borji, A.
\newblock Revisiting video saliency: A large-scale benchmark and a new model.
\newblock In \emph{Proc. of CVPR}, 2018.

\bibitem[Xiong et~al.(2015)Xiong, Sundaresan, and Jamieson]{xiong15tonetrack}
Xiong, J., Sundaresan, K., and Jamieson, K.
\newblock Tonetrack: Leveraging frequency-agile radios for time-based indoor
  wireless localization.
\newblock In \emph{Proc of MobiCom}, 2015.

\bibitem[Yu et~al.(2018)Yu, Xian, Chen, Liu, Liao, Madhavan, and
  Darrell]{bdd100k}
Yu, F., Xian, W., Chen, Y., Liu, F., Liao, M., Madhavan, V., and Darrell, T.
\newblock Bdd100k: A diverse driving video database with scalable annotation
  tooling.
\newblock \emph{CoRR}, abs/1805.04687, 2018.

\bibitem[Zeng et~al.(2014)Zeng, Ouyang, Wang, and Wang]{zeng2014deep}
Zeng, X., Ouyang, W., Wang, M., and Wang, X.
\newblock Deep learning of scene-specific classifier for pedestrian detection.
\newblock In \emph{Proc. of ECCV}, 2014.

\bibitem[Zhang et~al.(2012)Zhang, Islam, and Lu]{zhang2012review}
Zhang, D., Islam, M.~M., and Lu, G.
\newblock A review on automatic image annotation techniques.
\newblock \emph{Pattern Recognition}, 45, 2012.

\bibitem[Zhang et~al.(2016)Zhang, Lin, Liang, and He]{kaiming16}
Zhang, L., Lin, L., Liang, X., and He, K.
\newblock Is faster {R-CNN} doing well for pedestrian detection?
\newblock In \emph{Proc. of ECCV}, 2016.

\bibitem[Zhou et~al.(2017)Zhou, Brown, Snavely, and Lowe]{zhou2017unsupervised}
Zhou, T., Brown, M., Snavely, N., and Lowe, D.~G.
\newblock Unsupervised learning of depth and ego-motion from video.
\newblock In \emph{Proc. of {CVPR}}, 2017.

\bibitem[Zhou(2017)]{zhou2017brief}
Zhou, Z.
\newblock A brief introduction to weakly supervised learning.
\newblock \emph{National Science Review}, 2017.

\bibitem[Zhu et~al.(2012)Zhu, Vondrick, Ramanan, and Fowlkes]{zhu2012moredata}
Zhu, X., Vondrick, C., Ramanan, D., and Fowlkes, C.~C.
\newblock Do we need more training data or better models for object detection?
\newblock In \emph{Proc. of BMVC}, 2012.

\bibitem[Zitnick \& Dollar(2014)Zitnick and Dollar]{edgebox14}
Zitnick, L. and Dollar, P.
\newblock Edge boxes: Locating object proposals from edges.
\newblock In \emph{Proc. of ECCV}, 2014.

\end{thebibliography}
\balance
\begin{small}

\end{small}

\end{document}